\documentclass[11pt]{article}

\usepackage[preprint]{acl}

\usepackage{times}
\usepackage{latexsym}

\usepackage[T1]{fontenc}

\usepackage[utf8]{inputenc}

\usepackage{microtype}

\usepackage{inconsolata}

\usepackage{graphicx}

\usepackage{bm}
\usepackage{amsmath}
\usepackage{booktabs}
\usepackage{multirow}
\usepackage{array}
\usepackage[table]{xcolor}
\usepackage{makecell}
\usepackage[most]{tcolorbox}
\usepackage{amssymb}
\usepackage{xcolor}
\usepackage{hyperref}

\title{The Shape of Time: Video-Token Contrast for Temporal \\ Understanding in VideoLMs}

\author{Yumeng Shi \quad Quanyu Long \quad Yin Wu \quad Wenya Wang \\
       Nanyang Technological University \\
       \texttt{\{yumeng001, quanyu001, wuyi0023\}@e.ntu.edu.sg} \quad \texttt{wangwy@ntu.edu.sg}}

\begin{document}
\maketitle
\begin{abstract}
Seeing frames in order does not mean representing time.
Modern VideoLMs receive ordered video streams, yet their main supervision acts on generated text rather than video-token representations where event dynamics should first emerge.
This mismatch allows models to learn temporal answers from shortcuts such as objects, scenes, and language priors, without requiring internal video representations to capture event progression.
To address this, we propose VT-Contrast, a representation-level temporal counterfactual objective for VideoLMs.
Its design asks where temporal supervision should act and what temporal differences it should expose.
VT-Contrast supervises selected late-layer last-frame video tokens, where temporal information is expected to be integrated before language generation, and contrasts order-preserving views with same-video reordered counterfactuals graded by Kendall tau distance.
It requires no architectural changes, is compatible with diverse VideoLM training tasks, and improves overall performance across temporal understanding benchmarks. Our code is available at \textit{\href{https://github.com/ANDgate99/VT-Contrast}
{https://github.com/ANDgate99/VT-Contrast}}.
\end{abstract}

\section{Introduction}
Video Language Models (VideoLMs)~\cite{lin2024video, maaz2024video} are increasingly expected to reason about how events unfold over time~\cite{li2026universal, qin2025question}, rather than merely recognize visual content.
For example, two clips may contain nearly the same objects and scenes, while expressing opposite events under different temporal progressions: opening versus closing, entering versus leaving, or assembling versus disassembling.
Resolving such order-dependent distinctions requires models to capture the progression of visual states over time.

However, standard VideoLM training does not directly enforce this requirement.
While VideoLMs receive temporal information through ordered visual tokens, supervision is usually applied only to the generated text~\cite{zhang2023video, zhang2024llava}. 
The language modeling loss specifies the desired answer, yet does not require the resulting visual representations to reflect the event progression that supports it.
For order-sensitive videos, this is a weak constraint: clips with similar visual evidence but different temporal order may still lead to correct responses based on static cues or answer priors.
As a result, response-level supervision may leave video representations underconstrained along the dimension most relevant to event progression, even when the final responses are correct.
\begin{figure}
    \centering
    \includegraphics[width=\linewidth]{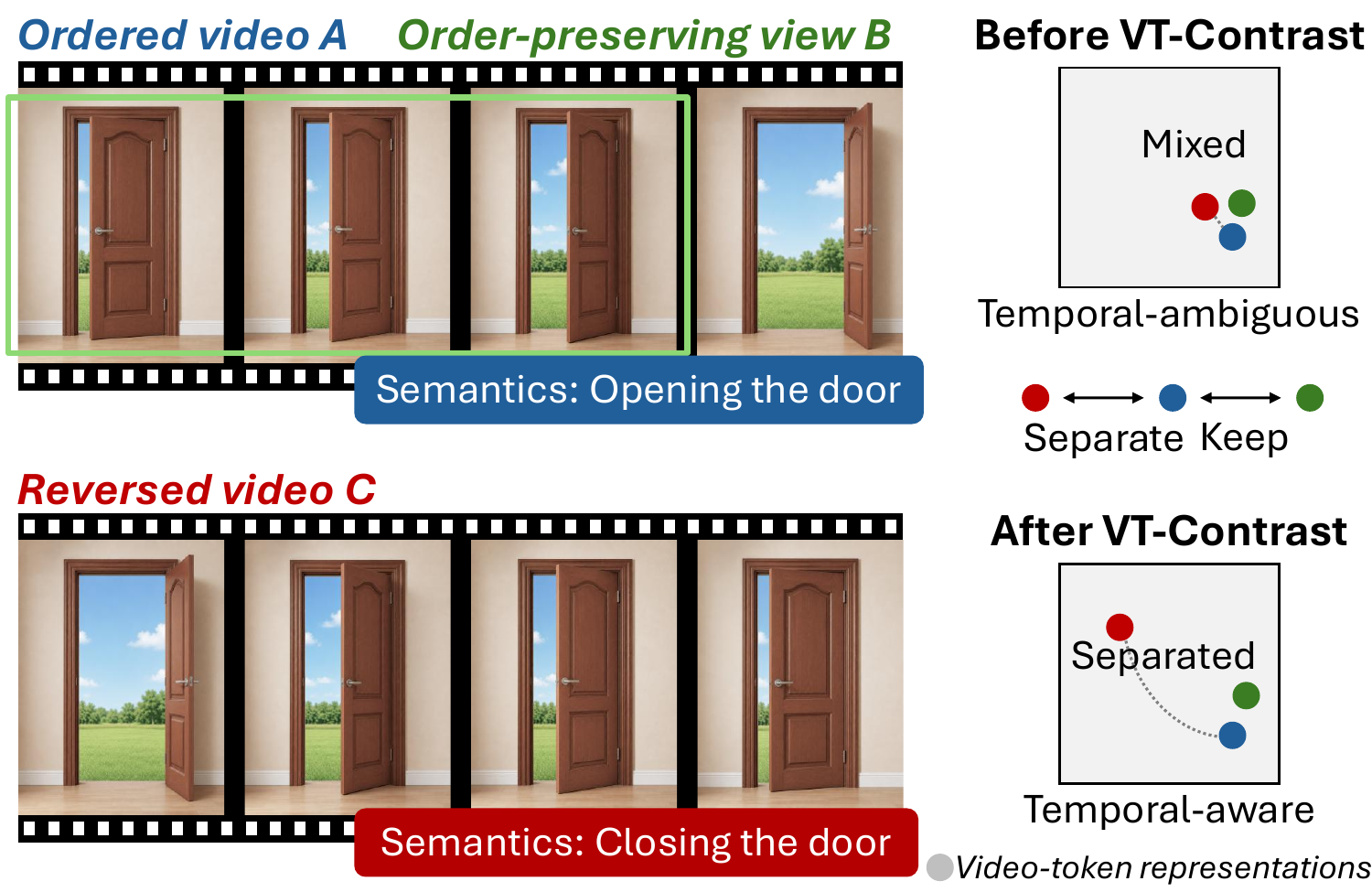}
    \caption{Motivation of VT-Contrast. Temporal order changes video semantics; VT-Contrast separates temporally corrupted negatives while keeping order-consistent positives close to the anchor in the video-token space.}
    \label{fig:intro}
\end{figure}

A more direct form of temporal supervision is therefore needed before response generation. Such supervision should make temporal order explicit in the internal video-token representation space, rather than relying solely on the final textual response. Prior approaches move in related directions but do not directly impose this constraint.
Temporal-aware architectures~\cite{hu2024enhancing} improve temporal aggregation in visual or intermediate modules, but do not directly supervise temporal-order representations after video tokens enter the language-model space.
Ordering-based auxiliary tasks~\cite{wu2025visual} expose models to shuffled segments, but still rely on task-level responses and can be less compatible with existing training objectives.
Reconstruction objectives~\cite{tong2022videomae} offer more direct visual supervision for representations, but often require extra branches and costly pixel-level targets.

A representation-level temporal signal should distinguish videos primarily by their temporal order rather than by their visual content.
We instantiate this idea by constructing multiple views from the same video: order-preserving views keep the original event progression, whereas reordered views retain the same visual evidence but change how the event unfolds.
As illustrated in Figure~\ref{fig:intro}, such order changes can alter event semantics, yet existing VideoLMs may still map the corresponding video-token representations close together.
We formulate these reordered views as temporal counterfactuals and propose \textbf{VT-Contrast}, a lightweight objective that separates them from order-preserving views in the video-token space.
This makes the learning signal depend on temporal order rather than video identity or static appearance cues.

Beyond this counterfactual construction, VT-Contrast must decide where the supervision should act and how strong the temporal signal should be.
For supervision placement, we apply the objective to selected late-layer last-frame video tokens, which are expected to aggregate information from preceding frames before language generation.
This choice is motivated by prior findings that temporal information emerges progressively inside VideoLM representations, rather than being uniformly available at all layers~\cite{zhang2025cross, shi2026causality}.
For temporal signal strength, we avoid treating all reorderings as equivalent.
We use Kendall tau distance~\cite{kendall1938new, diaconis1977spearman} to quantify pairwise order violations, enabling VT-Contrast to construct graded temporal counterfactuals that yield more effective contrastive supervision.

VT-Contrast is jointly optimized with the standard language modeling objective, requiring no architectural modification.
By directly shaping video-token geometry, it complements response-level supervision: the model is still trained to generate correct answers, while its internal video representations are encouraged to reflect event progression.
Experiments on multiple temporal understanding benchmarks show consistent overall gains, especially on order-sensitive tasks.
Our contributions are summarized as follows:
\begin{itemize}
    \item We introduce a representation-level temporal supervision perspective for VideoLMs, arguing that temporal order should be reflected in video-token representations rather than only supervised through textual responses.

    \item We propose VT-Contrast, a lightweight temporal counterfactual objective for VideoLM training that uses same-video temporally reordered sequences as negatives. It applies targeted supervision to selected-layer last-frame token representations and controls negative difficulty with Kendall tau distance.

    \item Evaluations on multiple temporal benchmarks show consistent overall improvements. VT-Contrast integrates into standard VideoLM training without changing the original task format, enabling flexible temporal enhancement across diverse video understanding scenarios.
\end{itemize}

\section{Related Works}
\subsection{Video Language Models}

Recent VideoLMs~\cite{zhang2024long, lin2024video} typically extend LLM-based multimodal understanding from images to videos by encoding frames or clips into visual tokens, projecting them into LLM feature space, and generating textual responses autoregressively~\cite{liu2023visual, maaz2024video}.
Representative systems include both closed-source models~\cite{comanici2025gemini, achiam2023gpt} and open-source models~\cite{bai2025qwen3, chen2024internvl, li2024llava}.
Despite differences in model scale, data, and training recipes, their training signal is primarily imposed on generated text tokens rather than on intermediate video-token representations~\cite{zhang2023video, zhang2024llava, feng2026video}.
Although language loss can update visual components, it does not directly constrain video-token representations to distinguish temporal order changes.

\begin{figure*}[t]
    \centering
    \includegraphics[width=\linewidth]{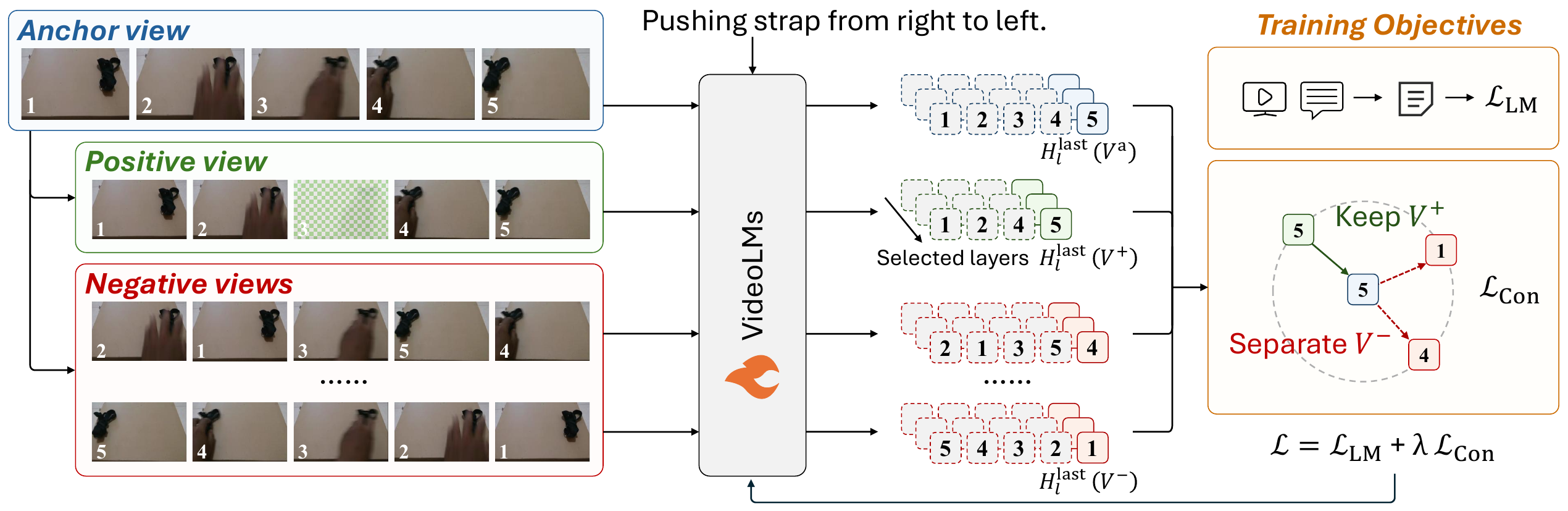}
    \caption{
Overview of VT-Contrast.
For each input video, we construct an anchor view, an order-preserving positive view, and temporal counterfactual negative views. 
A shared VideoLM processes these views to obtain selected video-token representations for contrastive supervision.
VT-Contrast keeps temporally consistent views close and separates negative views, and is jointly optimized with the standard language modeling loss.
}
    \label{fig:overview}
\end{figure*}

\subsection{Temporal Understanding}

Temporal understanding is fundamental to VideoLMs, as videos require models to reason over event order, motion direction, state changes, and other relations beyond static visual recognition~\cite{zhang2024vinoground, chandrasegaran2024hourvideo}.
Benchmarks such as ViLMA~\cite{DBLP:conf/iclr/KesenPDCAPCFGEE24}, TempCompass~\cite{liu2024tempcompass}, and TOMATO~\cite{shangguan2025tomato} show that current VideoLMs still struggle with fine-grained temporal semantics, especially when videos share similar visual content but differ in temporal order or event progression. 
To improve temporal understanding, prior work has explored temporal-aware architectures, temporal reasoning transfer, and auxiliary ordering tasks~\cite{hu2024enhancing, fateh2024video}.
For instance, T3~\cite{li2025temporal} transfers temporal reasoning from synthetic text tasks, while Visual Jigsaw~\cite{wu2025visual} trains models to recover shuffled visual order.
While these efforts highlight the importance of temporal modeling, directly shaping video-token representations for temporal sensitivity remains less explored.

\subsection{Contrastive Learning}

Contrastive learning is widely used for representation learning by pulling positive pairs together and pushing negative samples apart in feature space~\cite{he2020momentum, chen2020simple}. 
In vision-language learning, CLIP-style objectives align visual and textual representations~\cite{radford2021learning}, and recent studies further extend contrastive training to VLM-based multimodal embedding models~\cite{jiang2024vlm2vec, meng2025vlm2vec}.
Beyond image-text alignment, contrastive objectives have also been applied to video representation learning~\cite{pan2021videomoco, qian2021spatiotemporal}.
Some works further exploit temporal perturbations to enhance video representations~\cite{dave2022tclr, dorkenwald2022scvrl, DBLP:conf/iccv/JenniJ21}.
However, these methods mainly target general-purpose representations in standalone video encoders.
In contrast, VT-Contrast targets the underconstrained temporal structure of VideoLM video-token representations, directly separating temporally reordered videos in this internal representation space.

\section{Method}
We propose VT-Contrast, a lightweight temporal counterfactual learning method that enhances temporal order sensitivity in VideoLMs through video-token supervision. 
It provides direct representation-level supervision by contrasting temporally consistent views with reordered views from the same video. 
We then describe how VT-Contrast is integrated into VideoLM training.

\subsection{Problem Setup}

Given a common video-language training dataset, each sample consists of a video $V=\{v_1,\ldots,v_T\}$, a textual query $Q$, and a target response $Y=\{y_1,\ldots,y_N\}$, where $v_t$ denotes the $t$-th sampled frame.
A VideoLM processes $V$ to obtain video-token representations, which are used as visual context for the language model.
The response is predicted autoregressively as:
\begin{equation}
p_\theta(Y \mid V, Q)
=
\prod_{i=1}^{N}
p_\theta(y_i \mid y_{<i}, V, Q),
\end{equation}
where $p_\theta$ denotes the token distribution predicted by the VideoLM with parameters $\theta$.

Standard VideoLM training minimizes the negative log-likelihood of the target response:
\begin{equation}
\mathcal{L}_{\mathrm{LM}}
=
-\sum_{i=1}^{N}
\log p_\theta(y_i \mid y_{<i}, V, Q).
\end{equation}

Although this response-level objective aligns videos with textual answers, it supervises video-token representations only indirectly through the output loss.
Consequently, these representations are not explicitly encouraged to distinguish fine-grained temporal order changes, which can allow the model to rely on static visual cues or language priors when they are sufficient for response generation.
We therefore introduce an auxiliary contrastive objective based on temporal counterfactual views to directly supervise video-token representations during VideoLM training.

\subsection{Temporal Counterfactual Learning}
As illustrated in Figure~\ref{fig:overview}, VT-Contrast constructs temporally consistent and order-disrupted views from the same input video.
The order-disrupted views serve as temporal counterfactuals: they preserve the visual content of the original video while altering its event progression.
Instead of relying on textual responses for temporal supervision, our method contrasts these video views directly in the representation space.
This makes temporal order differences explicit in video-token representations without changing the response format.

\paragraph{Contrastive view construction.}

We instantiate this idea by constructing three views from each training video $V=\{v_1,\ldots,v_T\}$: an anchor view $V^a$, an order-consistent positive view $V^+$, and an order-disrupted negative view $V^-$.
The anchor defines the reference temporal progression, the positive preserves this progression under a mild sampling perturbation, and the counterfactual negative view changes the progression while keeping the sampled visual content fixed.
Together, these views allow the contrastive objective to compare temporal consistency against order disruption under controlled visual content.

\medskip
$\triangleright$ \textbf{Anchor:}
The anchor view $V^a$ is the reference chronological view. It is obtained by uniformly sampling $K$ frames from the original video while preserving their temporal order. Formally, let $i_1<i_2<\cdots<i_K$ denote the sampled frame indices; the anchor view is defined as:
\begin{equation}
V^a = \{v_{i_1}, v_{i_2}, \ldots, v_{i_K}\}.
\end{equation}

\medskip
$\triangleright$ \textbf{Positive:}
The positive view $V^+$ is an order-consistent counterpart to the anchor view.
It preserves the temporal progression and main visual content of $V^a$ while introducing a mild sampling variation.
In this work, we instantiate this perturbation by randomly dropping one frame from $V^a$.
Let $d \in \{1,\ldots,K\}$ denote the dropped position; the positive view is defined as:
\begin{equation}
V^+ =
\{v_{i_1}, \ldots, v_{i_{d-1}}, v_{i_{d+1}}, \ldots, v_{i_K}\}.
\end{equation}

Since the remaining frames preserve the original chronological order, $V^a$ and $V^+$ are treated as a positive pair.

\medskip
$\triangleright$ \textbf{Negative:}
The negative view $V^-$ is designed to isolate temporal order as the main contrasting factor.
It reuses the same sampled frames as the anchor view, so that objects, scenes, and local visual patterns are largely preserved while the chronological order is disrupted.
This same-video construction reduces shortcuts based on video identity or static appearance, making temporal order the primary difference between $V^a$ and $V^-$.

Specifically, we apply a non-identity temporal permutation $\pi$ to the anchor sequence:
\begin{equation}
V^- = \{v_{i_{\pi(1)}}, v_{i_{\pi(2)}}, \ldots, v_{i_{\pi(K)}}\},
\quad \pi \neq \mathrm{id}.
\end{equation}

Thus, $V^-$ serves as a temporal counterfactual negative that differs from $V^a$ in frame order.

\paragraph{Contrastive objective.}

To realize temporal counterfactual learning, we train the anchor representation to stay close to its order-consistent positive view and away from same-video counterfactual negatives.
In this way, the contrastive objective turns temporal order changes into explicit constraints on video-token representations.
Specifically, we adopt an InfoNCE objective~\cite{oord2018representation}.
For each anchor view, we construct $M$ counterfactual negative views $\{V_m^-\}_{m=1}^{M}$.
Let $r(U)$ denote the video-token representation extracted from any constructed view $U$.
We compute the positive and negative similarity scores as:
\begin{equation}
\begin{aligned}
s^+ &= \mathrm{sim}\big(r(V^a),r(V^+)\big), \\
s_m^- &= \mathrm{sim}\big(r(V^a),r(V_m^-)\big).
\end{aligned}
\end{equation}
The contrastive loss is defined as:
\begin{equation}
\mathcal{L}_{\mathrm{con}}
=
-\log
\frac{\exp(s^+/\tau_c)}
{\exp(s^+/\tau_c)
+
\sum_{m=1}^{M}\exp(s_m^-/\tau_c)} ,
\end{equation}
where $\mathrm{sim}(\cdot,\cdot)$ denotes cosine similarity and $\tau_c$ is the contrastive temperature.

\subsection{Emergence-aware Temporal Supervision}

The temporal counterfactual objective should not be applied blindly to all video tokens or all reordered views.
We therefore design the supervision to be both targeted and graded, so that the representation constraint focuses on informative temporal features while keeping the counterfactual comparisons at an appropriate difficulty.

\begin{figure}
    \centering
    \includegraphics[width=\linewidth]{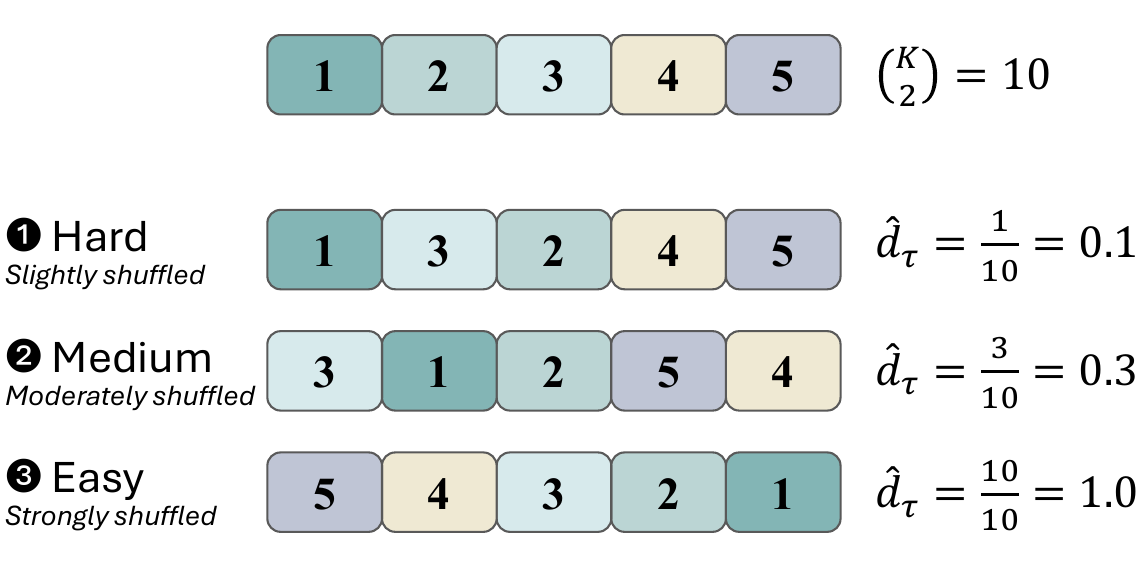}
    \caption{
Kendall-tau-based grading of temporal counterfactual negatives.
Smaller $\hat{d}_{\tau}$ yields harder counterfactuals closer to the original order, while larger $\hat{d}_{\tau}$ indicates stronger order disruption and easier contrasts.
}
    \label{fig:kendall}
\end{figure}

\paragraph{Targeted video-token supervision.}

We now describe how the contrastive representation $r(U)$ is extracted for each constructed view.
Each view is processed by the shared VideoLM to obtain layer-wise video-token features.
Motivated by the observation that different layers capture different stages of multimodal information flow and last-frame features can aggregate temporal cues across the video~\citep{zhang2025cross, shi2026causality}, we use last-frame representations from selected layers $\mathcal{S}$ for contrastive supervision.
This targeted design avoids dense constraints over all layers and video tokens while focusing on compact, temporally informative features.

Specifically, for each view $U$ of the original video $V$ and selected layer $l \in \mathcal{S}$, we take the last-frame video-token features, denoted as $H_l^{\mathrm{last}}(U)$.
We first apply mean pooling over the last-frame tokens at each selected layer and then average the resulting features across selected layers:
\begin{equation}
r(U)
=
\frac{1}{|\mathcal{S}|}
\sum_{l \in \mathcal{S}}
\mathrm{MeanPool}\left(H_l^{\mathrm{last}}(U)\right),
\end{equation}
where $\mathrm{MeanPool}(\cdot)$ averages the last-frame video tokens into a single feature vector.
Since different VideoLM backbones vary in depth and representation structure, the selected layer set $\mathcal{S}$ is model-dependent.
We specify the exact layers in the implementation details.

\begin{table*}[t]
\centering
\fontsize{8pt}{9pt}\selectfont
\setlength{\tabcolsep}{1.6pt}
\renewcommand{\arraystretch}{1.03}

\newcolumntype{F}{>{\centering\arraybackslash}p{0.055\textwidth}}
\newcolumntype{C}{>{\raggedright\arraybackslash}p{0.215\textwidth}}
\newcolumntype{A}{>{\centering\arraybackslash}p{0.052\textwidth}}
\newcolumntype{G}{>{\raggedright\arraybackslash}p{0.03\textwidth}}

\newcommand{\posgain}[1]{{\hspace{-0.2em}\fontsize{5pt}{5pt}\selectfont\itshape\textcolor{green!50!black}{#1}}}
\newcommand{\neggain}[1]{{\hspace{-0.2em}\fontsize{5pt}{5pt}\selectfont\itshape\textcolor{red!70!black}{#1}}}

\definecolor{oursgray}{gray}{0.95}
\newcommand{\oursrow}{\rowcolor{oursgray}}

\definecolor{oursrulegray}{gray}{0.85}
\newcommand{\oursline}{\arrayrulecolor{oursrulegray}}

\begin{tabular}{
F C
A@{\hspace{0.0em}}G
A@{\hspace{0.0em}}G
A@{\hspace{0.0em}}G
A@{\hspace{0.0em}}G
A@{\hspace{0.0em}}G
A@{\hspace{0.0em}}G
A@{\hspace{0.0em}}G
A@{\hspace{0.0em}}G
}
\toprule
\multirow{2}{*}{\textbf{Frames}} 
& \multirow{2}{*}{\textbf{Model}} 
& {\multirow{2}{*}{\hspace{-0.83em}\textbf{TOMATO}}}
&
& \multicolumn{8}{c}{\textbf{TempCompass}} 
& \multicolumn{6}{c}{\textbf{Vinoground}} \\
\cmidrule(lr){5-12} \cmidrule(lr){13-18}
& 
& \multicolumn{2}{c}{}
& \textbf{Y/N} & 
& \textbf{MCQ} & 
& \makebox[0pt][l]{\hspace{-1.72em}\textbf{Caption}} & 
& \textbf{Match} & 
& \textbf{Text} & 
& \textbf{Video} & 
& \textbf{Group} & \\
\midrule

\multirow{9}{*}{\textbf{8}}
& LLaVA-OneVision1.5-4B 
& 21.09 & 
& 62.86 & 
& 56.71 & 
& 50.00 & 
& 68.26 & 
& 20.60 & 
& 23.80 & 
& 3.60 & \\
& InternVL3.5-4B 
& 28.71 & 
& 72.08 & 
& 68.73 & 
& 60.93 & 
& 79.44 & 
& 48.20 & 
& 28.20 & 
& 16.00 & \\
& Visual Jigsaw 
& 26.01 & 
& 73.50 & 
& 71.77 & 
& 61.93 & 
& 79.51 & 
& 44.80 & 
& 25.00 & 
& 13.20 & \\

\oursline
\cmidrule(lr){2-18}
\arrayrulecolor{black}

& Qwen3.5-0.8B 
& 26.01 & 
& 63.64 & 
& 58.80 & 
& 48.35 & 
& 70.39 & 
& 26.60 & 
& 22.00 & 
& 7.00 & \\
& Qwen3.5-2B 
& 25.74 & 
& 71.02 & 
& 67.34 & 
& 56.64 & 
& 77.45 & 
& 34.00 & 
& 20.40 & 
& 7.60 & \\
& Qwen3.5-4B 
& 33.42 & 
& 77.62 & 
& 71.90 & 
& 49.50 & 
& 83.77 & 
& 50.00 & 
& 31.80 & 
& 19.60 & \\

\oursline
\cmidrule(lr){2-18}
\arrayrulecolor{black}

\oursrow
\cellcolor{white} & Ours-0.8B
& \textbf{27.29} & \posgain{+1.28}
& \textbf{64.98} & \posgain{+1.34}
& \textbf{59.49} & \posgain{+0.69}
& \textbf{50.05} & \posgain{+1.70}
& 70.26 & \neggain{-0.13}
& \textbf{29.40} & \posgain{+2.80}
& \textbf{22.20} & \posgain{+0.20}
& \textbf{7.20} & \posgain{+0.20} \\

\oursrow
\cellcolor{white} & Ours-2B
& \textbf{31.47} & \posgain{+5.73}
& \textbf{74.97} & \posgain{+3.95}
& \textbf{70.00} & \posgain{+2.66}
& \textbf{64.17} & \posgain{+7.53}
& \textbf{80.17} & \posgain{+2.72}
& \textbf{41.20} & \posgain{+7.20}
& \textbf{25.00} & \posgain{+4.60}
& \textbf{11.40} & \posgain{+3.80} \\

\oursrow
\cellcolor{white} & Ours-4B
& \textbf{36.46} & \posgain{+3.04}
& \textbf{79.33} & \posgain{+1.71}
& \textbf{74.94} & \posgain{+3.04}
& \textbf{50.80} & \posgain{+1.30}
& \textbf{85.30} & \posgain{+1.53}
& \textbf{57.20} & \posgain{+7.20}
& 29.00 & \neggain{-2.80}
& \textbf{20.00} & \posgain{+0.40} \\

\midrule

\multirow{9}{*}{\textbf{16}}
& LLaVA-OneVision1.5-4B 
& 20.55 & 
& 62.25 & 
& 57.47 & 
& 51.15 & 
& 69.33 & 
& 20.00 & 
& 23.80 & 
& 4.80 & \\
& InternVL3.5-4B 
& 30.12 & 
& 72.89 & 
& 70.70 & 
& 61.68 & 
& 80.44 & 
& 53.60 & 
& 30.80 & 
& 16.80 & \\
& Visual Jigsaw 
& 25.13 & 
& 74.60 & 
& 72.78 & 
& 62.72 & 
& 80.64 & 
& 47.20 & 
& 28.00 & 
& 15.40 & \\

\oursline
\cmidrule(lr){2-18}
\arrayrulecolor{black}

& Qwen3.5-0.8B 
& 25.81 & 
& 64.04 & 
& 59.94 & 
& 49.75 & 
& 70.73 & 
& 27.60 & 
& 21.20 & 
& 5.80 & \\
& Qwen3.5-2B 
& 26.55 & 
& 72.48 & 
& 69.18 & 
& 58.83 & 
& 78.84 & 
& 37.40 & 
& 23.00 & 
& 9.60 & \\
& Qwen3.5-4B 
& 35.58 & 
& 77.66 & 
& 74.68 & 
& 50.90 & 
& 84.56 & 
& 57.80 & 
& 33.40 & 
& 22.40 & \\

\oursline
\cmidrule(lr){2-18}
\arrayrulecolor{black}

\oursrow
\cellcolor{white} & Ours-0.8B
& \textbf{27.63} & \posgain{+1.82}
& \textbf{65.27} & \posgain{+1.23}
& \textbf{60.95} & \posgain{+1.01}
& \textbf{50.15} & \posgain{+0.40}
& \textbf{71.39} & \posgain{+0.66}
& \textbf{31.00} & \posgain{+3.40}
& \textbf{22.60} & \posgain{+1.40}
& \textbf{8.80} & \posgain{+3.00} \\

\oursrow
\cellcolor{white} & Ours-2B
& \textbf{31.20} & \posgain{+4.65}
& \textbf{76.07} & \posgain{+3.59}
& \textbf{71.46} & \posgain{+2.28}
& \textbf{65.77} & \posgain{+6.94}
& \textbf{82.24} & \posgain{+3.40}
& \textbf{43.40} & \posgain{+6.00}
& \textbf{24.80} & \posgain{+1.80}
& \textbf{12.80} & \posgain{+3.20} \\

\oursrow
\cellcolor{white} & Ours-4B
& \textbf{37.67} & \posgain{+2.09}
& \textbf{79.49} & \posgain{+1.83}
& \textbf{76.20} & \posgain{+1.52}
& 50.55 & \neggain{-0.35}
& \textbf{85.63} & \posgain{+1.07}
& \textbf{59.80} & \posgain{+2.00}
& 31.60 & \neggain{-1.80}
& 21.00 & \neggain{-1.40} \\

\midrule

\multirow{9}{*}{\textbf{32}}
& LLaVA-OneVision1.5-4B 
& 20.49 & 
& 62.49 & 
& 56.71 & 
& 49.85 & 
& 68.80 & 
& 19.80 & 
& 23.40 & 
& 4.80 & \\
& InternVL3.5-4B 
& 29.65 & 
& 73.46 & 
& 70.76 & 
& 62.03 & 
& 80.71 & 
& 52.80 & 
& 32.60 & 
& 19.40 & \\
& Visual Jigsaw 
& 26.48 & 
& 74.52 & 
& 72.78 & 
& 61.93 & 
& 80.24 & 
& 48.80 & 
& 30.80 & 
& 18.60 & \\

\oursline
\cmidrule(lr){2-18}
\arrayrulecolor{black}

& Qwen3.5-0.8B 
& 26.08 & 
& 64.57 & 
& 60.89 & 
& 51.50 & 
& 71.99 & 
& 28.00 & 
& 20.00 & 
& 5.80 & \\
& Qwen3.5-2B 
& 27.02 & 
& 72.97 & 
& 69.43 & 
& 58.13 & 
& 79.57 & 
& 38.20 & 
& 22.60 & 
& 9.20 & \\
& Qwen3.5-4B 
& 36.39 & 
& 78.60 & 
& 74.62 & 
& 48.65 & 
& 84.96 & 
& 59.20 & 
& 33.80 & 
& 21.80 & \\

\oursline
\cmidrule(lr){2-18}
\arrayrulecolor{black}

\oursrow
\cellcolor{white} & Ours-0.8B
& \textbf{28.17} & \posgain{+2.09}
& \textbf{66.08} & \posgain{+1.51}
& \textbf{62.22} & \posgain{+1.33}
& 50.10 & \neggain{-1.40}
& \textbf{72.12} & \posgain{+0.13}
& \textbf{30.60} & \posgain{+2.60}
& \textbf{25.60} & \posgain{+5.60}
& \textbf{8.80} & \posgain{+3.00} \\

\oursrow
\cellcolor{white} & Ours-2B
& \textbf{30.93} & \posgain{+3.91}
& \textbf{75.95} & \posgain{+2.98}
& \textbf{71.71} & \posgain{+2.28}
& \textbf{65.52} & \posgain{+7.39}
& \textbf{82.44} & \posgain{+2.87}
& \textbf{44.60} & \posgain{+6.40}
& \textbf{25.40} & \posgain{+2.80}
& \textbf{13.20} & \posgain{+4.00} \\

\oursrow
\cellcolor{white} & Ours-4B
& \textbf{38.41} & \posgain{+2.02}
& \textbf{79.54} & \posgain{+0.94}
& \textbf{76.58} & \posgain{+1.96}
& \textbf{50.75} & \posgain{+2.10}
& \textbf{86.09} & \posgain{+1.13}
& \textbf{60.40} & \posgain{+1.20}
& \textbf{34.00} & \posgain{+0.20}
& \textbf{22.60} & \posgain{+0.80} \\

\bottomrule
\end{tabular}
\caption{
Main results on TOMATO, TempCompass, and Vinoground under different frame settings.
Our 4B model performs better than recent 4B MLLMs on most benchmarks.
Our method also improves over the corresponding Qwen3.5 models in most settings, showing that VT-Contrast strengthens temporal understanding without architectural changes.
Small colored numbers denote absolute gains over the corresponding Qwen3.5 model.
}
\label{tab:main_results}
\end{table*}

\paragraph{Graded temporal counterfactuals.}
Temporal counterfactuals should differ from the anchor in temporal order, but the degree of order violation affects the learning signal.
Smaller violations create harder fine-grained contrasts, while larger violations produce more explicit but easier counterfactuals.
Therefore, we use Kendall tau distance~\cite{kendall1938new, diaconis1977spearman} to grade the strength of order violation, allowing us to control the signal and construct effective temporal contrasts, as illustrated in Figure~\ref{fig:kendall}.

Given a permutation $\pi_m$ over the $K$ frames of the anchor view, the Kendall tau distance counts the number of pairwise order inversions:
\begin{equation}
d_{\tau}(\pi_m)
=
\sum_{1 \le i < j \le K}
\mathbb{I}\!\left[\pi_m(i)>\pi_m(j)\right],
\end{equation}
where $\mathbb{I}[\cdot]$ is the indicator function.
Since the maximum number of inversions is $K(K-1)/2$, we normalize the distance as:
\begin{equation}
\hat{d}_{\tau}(\pi_m)
=
\frac{2d_{\tau}(\pi_m)}{K(K-1)}.
\end{equation}

The normalized distance $\hat{d}_{\tau}(\pi_m) \in [0,1]$ measures how strongly the counterfactual violates the original temporal order.
Smaller values indicate that the permutation remains close to the original order and is harder to distinguish, whereas larger values indicate stronger order disruption.

During training, we sample each counterfactual permutation $\pi_m$ from a controlled range:
\begin{equation}
\delta_{\min}
<
\hat{d}_{\tau}(\pi_m)
\leq
\delta_{\max},
\end{equation}
where $\delta_{\min}$ and $\delta_{\max}$ define the allowed degree of order violation.
If the number of valid permutations in the range is insufficient, we repeatedly sample with replacement until obtaining the required $M$ negatives.
In our implementation, we use $0 < \hat{d}_{\tau}(\pi_m) \leq 0.5$.
This favors counterfactuals that introduce order violations while remaining relatively close to the original sequence, making them harder to distinguish.

\subsection{Optimization Objective}

We keep the standard language modeling objective: it provides semantic supervision for the original temporally ordered video-query pair and preserves the model's generation ability.
VT-Contrast is added as an auxiliary objective, providing direct temporal supervision on video-token representations.
The final training loss is:
\begin{equation}
\mathcal{L}
=
\mathcal{L}_{\mathrm{LM}}
+
\lambda \mathcal{L}_{\mathrm{con}},
\end{equation}
where $\lambda$ controls the balance between language modeling and temporal contrastive supervision.
In all experiments, we set $\lambda=0.1$, which enables stable joint optimization.

\section{Experiments}
In this section, we evaluate the proposed method through extensive experiments. 
We first describe the experimental setup, present the main results, and then analyze key design choices and learned representations.
We also provide qualitative analysis in Appendix~\ref{sec:appendix_qualitative}.

\newcommand{\metricw}{0.063\textwidth}

\definecolor{oursgray}{gray}{0.95}
\newcommand{\oursrow}{\rowcolor{oursgray}}
\newcolumntype{L}[1]{>{\raggedright\arraybackslash}p{#1}}
\newcolumntype{C}[1]{>{\centering\arraybackslash}p{#1}}
\newcommand{\gain}[1]{{\fontsize{8pt}{9pt}\selectfont\textcolor{green!50!black}{\textit{#1}}}}

\begin{table*}[t]
\centering
\fontsize{8pt}{9pt}\selectfont
\renewcommand{\arraystretch}{1.1}
\begin{tabular}{
L{0.104\textwidth}
C{0.05\textwidth}
C{\metricw}
C{\metricw}
C{\metricw}
C{\metricw}
C{\metricw}
C{\metricw}
C{\metricw}
C{\metricw}
}
\toprule
\multirow{2}{*}{\textbf{Setting}}
& \multirow{2}{*}{\textbf{$\lambda$}}
& \multirow{2}{*}{\textbf{TOMATO}}
& \multicolumn{4}{c}{\textbf{TempCompass}}
& \multicolumn{3}{c}{\textbf{Vinoground}} \\
\cmidrule(lr){4-7} \cmidrule(lr){8-10}
& &
& \textbf{Y/N}
& \textbf{MCQ}
& \textbf{Caption}
& \textbf{Match}
& \textbf{Text}
& \textbf{Video}
& \textbf{Group} \\
\midrule
LM only
& 0
& 29.45
& 72.85
& 68.86
& 62.72
& 78.84
& 36.40
& 22.00
& 9.20 \\

+ VT-Contrast
& 0.1
& \textbf{31.47}
& \textbf{74.97}
& \textbf{70.00}
& \textbf{64.17}
& \textbf{80.17}
& \textbf{41.20}
& \textbf{25.00}
& \textbf{11.40} \\

\oursrow
{\gain{Gain}}
& {\scriptsize --}
& \gain{+2.02}
& \gain{+2.12}
& \gain{+1.14}
& \gain{+1.45}
& \gain{+1.33}
& \gain{+4.80}
& \gain{+3.00}
& \gain{+2.20} \\
\bottomrule
\end{tabular}
\caption{
Effect of the temporal counterfactual objective.
Compared with LM-only training, VT-Contrast consistently improves across benchmarks, showing the benefit of direct temporal supervision on video-token representations.
Gains denote absolute improvements over the LM-only baseline.
}
\label{tab:ab_loss}
\end{table*}

\subsection{Experimental Settings}

\paragraph{Baselines.}
We compare models trained with our method against four representative baselines.
InternVL3.5~\cite{wang2025internvl3} and LLaVA-OneVision1.5~\cite{an2025llava} are recent open-source multimodal large language models; we report their 4B variants.
Visual Jigsaw~\cite{wu2025visual} is a 7B model tailored for temporal understanding and trained for order prediction with GRPO~\cite{shao2024deepseekmath}.
Finally, we include Qwen3.5~\cite{qwen3.5}, the base model used in our training, and evaluate its 0.8B, 2B, and 4B variants to study performance across scales.

\begin{table*}[t]
\centering
\fontsize{8pt}{9pt}\selectfont
\renewcommand{\arraystretch}{1.1}
\begin{tabular}{
L{0.18\textwidth}
C{\metricw}
C{\metricw}
C{\metricw}
C{\metricw}
C{\metricw}
C{\metricw}
C{\metricw}
C{\metricw}
}
\toprule
\multirow{2}{*}{\textbf{Token Representation}}
& \multirow{2}{*}{\textbf{TOMATO}}
& \multicolumn{4}{c}{\textbf{TempCompass}}
& \multicolumn{3}{c}{\textbf{Vinoground}} \\
\cmidrule(lr){3-6} \cmidrule(lr){7-9}
& 
& \textbf{Y/N}
& \textbf{MCQ}
& \textbf{Caption}
& \textbf{Match}
& \textbf{Text}
& \textbf{Video}
& \textbf{Group} \\
\midrule
Last query token
& \textbf{31.81}
& 71.50
& 68.86
& 59.23
& 77.45
& 33.40
& \textbf{25.80}
& \underline{9.80} \\

All frame tokens
& 30.26
& \underline{73.99}
& \textbf{70.19}
& \underline{62.92}
& \underline{79.77}
& \underline{40.20}
& 21.40
& \underline{9.80} \\

\oursrow
Last-frame tokens
& \underline{31.47}
& \textbf{74.97}
& \underline{70.00}
& \textbf{64.17}
& \textbf{80.17}
& \textbf{41.20}
& \underline{25.00}
& \textbf{11.40} \\
\bottomrule
\end{tabular}
\caption{
Effect of token representation for temporal supervision.
Last-frame video tokens achieve the best overall performance, suggesting that targeted supervision on temporally aggregated video-token representations is more effective for VT-Contrast.
Bold and underlined numbers indicate the best and second-best results, respectively.
}
\label{tab:ab_token}
\end{table*}

\paragraph{Datasets.}
For training, we use the Something-Something V2 (SSv2)~\cite{goyal2017something} training split, a large-scale dataset for fine-grained human-object action understanding.

For evaluation, we use three benchmarks focused on temporal capability: TOMATO~\cite{shangguan2025tomato}, TempCompass~\cite{liu2024tempcompass}, and Vinoground~\cite{zhang2024vinoground}.
TOMATO comprises 1,484 questions over 1,417 videos and evaluates temporal reasoning over event progression and state changes, while TempCompass contains 7,540 task instructions covering action, speed, direction, and event order.
Vinoground comprises 2,000 evaluation questions derived from 500 counterfactual pairs and focuses on fine-grained video-caption alignment under subtle temporal variations.
We report accuracy across all benchmarks, where higher values indicate better temporal understanding.

\paragraph{Implementation details.}
Our models are initialized from Qwen3.5~\cite{qwen3.5}, a recent state-of-the-art open-source model, at three scales: 0.8B, 2B, and 4B.
We fine-tune them on the SSv2~\cite{goyal2017something} training split under a video question answering formulation, using ``What action is happening in the video?'' as the prompt and the action label as the target response.
We jointly optimize the language modeling loss and VT-Contrast loss, setting $\lambda=0.1$ throughout the main text.
For VT-Contrast, we set the contrastive temperature to $\tau_c=0.10$, sample 16 negatives with replacement from permutations with Kendall tau distance in $(0,0.5]$, uniformly sample 8 frames per video, and use last-frame video tokens.
We apply VT-Contrast to layers 21--24 out of 24 layers for the 0.8B and 2B models, and to layers 25--28 out of 32 layers for the 4B model.
All models are trained for 2,250 steps using the same global batch size of 16 and a learning rate of $1\times10^{-5}$.

During evaluation, we use the lmms-eval framework~\cite{zhang2025lmms}, disable thinking mode, and adopt greedy decoding for deterministic results.
To reduce formatting-related evaluation errors, we use prompts with standardized answer formats; details are provided in Appendix~\ref{sec:appendix_prompt}.

\begin{table*}[t]
\centering
\fontsize{8pt}{9pt}\selectfont
\renewcommand{\arraystretch}{1.1}
\begin{tabular}{
L{0.18\textwidth}
C{\metricw}
C{\metricw}
C{\metricw}
C{\metricw}
C{\metricw}
C{\metricw}
C{\metricw}
C{\metricw}
}
\toprule
\multirow{2}{*}{\textbf{Counterfactual Difficulty}}
& \multirow{2}{*}{\textbf{TOMATO}}
& \multicolumn{4}{c}{\textbf{TempCompass}}
& \multicolumn{3}{c}{\textbf{Vinoground}} \\
\cmidrule(lr){3-6} \cmidrule(lr){7-9}
& 
& \textbf{Y/N}
& \textbf{MCQ}
& \textbf{Caption}
& \textbf{Match}
& \textbf{Text}
& \textbf{Video}
& \textbf{Group} \\
\midrule
Full negatives
& \underline{30.86}
& 73.62
& \underline{69.94}
& 63.22
& 79.71
& \underline{39.20}
& \textbf{26.60}
& 10.80 \\

Simple negatives
& 30.80
& \underline{73.75}
& 68.92
& \underline{63.82}
& \underline{80.04}
& \underline{39.20}
& 24.20
& \underline{11.00} \\

\oursrow
Hard negatives
& \textbf{31.47}
& \textbf{74.97}
& \textbf{70.00}
& \textbf{64.17}
& \textbf{80.17}
& \textbf{41.20}
& \underline{25.00}
& \textbf{11.40} \\
\bottomrule
\end{tabular}
\caption{
Effect of temporal counterfactual difficulty.
We compare full $(0,1]$, simple $(0.5,1]$, and hard $(0,0.5]$ negatives based on Kendall-tau distance.
The best performance from hard counterfactuals shows that Kendall-tau-based grading helps select more useful temporal counterfactuals.
}
\label{tab:ab_negative_hardness}
\end{table*}
\begin{figure*}
    \centering
    \includegraphics[width=1\linewidth]{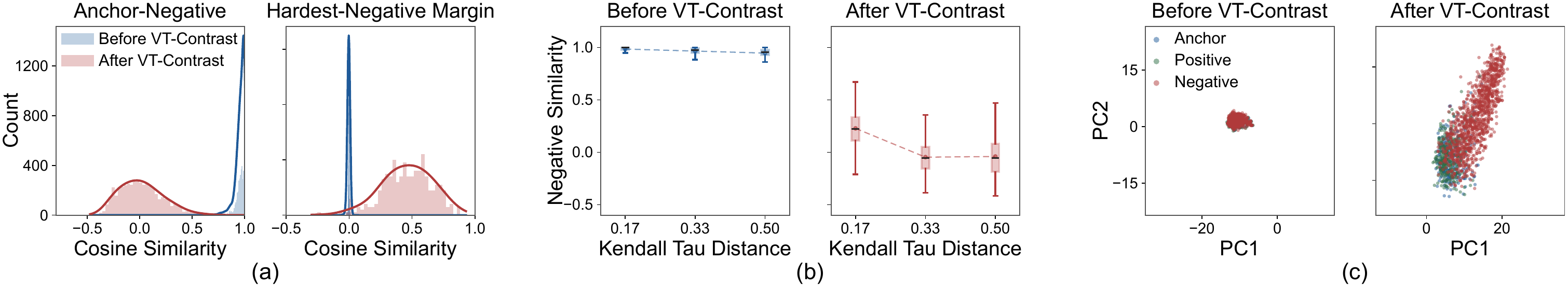}
\caption{
Representation analysis of VT-Contrast.
(a) Similarity distributions: VT-Contrast separates hardest negatives from positives while keeping anchor-positive similarities relatively high.
(b) Distance-similarity comparison: negative similarity decreases more clearly with Kendall-tau distance after training.
(c) PCA visualization: anchor, positive, and negative representations become more distinguishable after training.
}
\label{fig:rep_analysis}
\label{fig:rep_analysis}
    \label{fig:placeholder}
\end{figure*}

\subsection{Main Results}

Table~\ref{tab:main_results} reports the main results under different frame budgets.
Following lmms-eval, frames are first sampled at the default FPS and capped by the specified frame budget, so the reported frame setting is an upper bound and some videos may use fewer frames.
Compared with recent open-source VideoLMs, our 4B model achieves stronger results on most temporal benchmarks, with a particularly clear advantage on TOMATO.

Compared with the corresponding Qwen3.5 base models, our method brings consistent gains across most model scales and frame settings.
The gains are particularly notable for the 2B model: under the 8-frame setting, it improves TOMATO by 5.73\% and TempCompass Y/N by 3.95\%, showing clear benefits on order-sensitive tasks.
The 0.8B and 4B models also show clear improvements, suggesting that the proposed objective remains effective across model scales, with further room for scale-specific tuning.
At the same time, the TempCompass Caption results suggest that some subtasks remain partly dependent on the base model capability under greedy decoding; in our results, the 2B model performs better than the 4B model on Caption and also obtains a larger gain.

\subsection{Ablation Studies}

We ablate the key designs of VT-Contrast, including the temporal counterfactual objective, targeted video-token representations, and counterfactual difficulty control.
All experiments use the 2B model under the 8-frame setting.
Additional analyses of layer selection, the number of negatives, and temperature are provided in Appendices~\ref{sec:appendix_layer}, \ref{sec:appendix_num}, and \ref{sec:appendix_temperature}, respectively.

\paragraph{Temporal counterfactual objective.}
Table~\ref{tab:ab_loss} ablates the temporal counterfactual objective.
Compared with LM-only training, VT-Contrast improves performance across the three benchmarks, with gains on every metric.
These consistent gains suggest that the objective provides useful supervision beyond standard language modeling, with larger improvements on TOMATO and Vinoground Text.
We use $\lambda=0.1$ by default, with additional loss-weight experiments in Appendix~\ref{sec:appendix_weight}.

\paragraph{Targeted video-token selection.}
Table~\ref{tab:ab_token} studies where to apply VT-Contrast.
Compared with supervising the last query token used for answer generation, supervising video-token representations achieves substantially better results on TempCompass and Vinoground Group.
This highlights the importance of applying temporal counterfactual supervision directly in the video-token space.
Among video-token choices, last-frame tokens outperform full-frame tokens overall, suggesting that the final-frame representations provide a compact target that better aggregates temporal information.

\paragraph{Temporal counterfactual difficulty.}
Table~\ref{tab:ab_negative_hardness} ablates Kendall-tau-based negative selection.
Hard negatives achieve the best overall performance across most benchmarks.
Their advantage over full negatives suggests that a focused hard subset is better than mixing widely varying difficulties, while their advantage over simple negatives shows the benefit of more challenging perturbations.
Together, these results demonstrate the importance of Kendall-tau-based grading for selecting effective temporal counterfactuals.

\subsection{Representation Analysis}

To understand how VT-Contrast improves temporal understanding, we analyze video-token representations on 300 SSv2 validation samples.
As shown in Figure~\ref{fig:rep_analysis}(a), before training, original videos and their reordered variants have very high cosine similarity and are mapped close in representation space.
This supports our motivation that standard VideoLM representations are insufficiently sensitive to temporal order perturbations.

After training, reordered variants are less concentrated around the original representation, and the hardest-negative margin increases.
Figure~\ref{fig:rep_analysis}(b) relates anchor-negative similarity to Kendall-tau distance, showing a clearer decrease with stronger temporal perturbations.
Figure~\ref{fig:rep_analysis}(c) shows PCA, where negative representations become more distinguishable from anchor and positive views.
Overall, these results show that same-video temporal counterfactuals provide effective supervision for order-sensitive video-token representations.

\section{Conclusion}

We propose VT-Contrast, a representation-level temporal supervision method for VideoLMs. 
The central idea is to shift temporal learning from response space to video-token space by contrasting order-preserving views with same-video reordered counterfactuals.
Applied to targeted last-frame representations, VT-Contrast uses Kendall tau distance to grade the temporal disruption of counterfactuals, making temporal order explicit in internal representations.
Experiments across temporal benchmarks show consistent overall gains, suggesting that shaping video-token representations is promising for improving temporal reasoning in VideoLMs.

\section*{Limitations}

Although VT-Contrast shows consistent improvements, our validation is limited by computational resources to moderate-scale supervised fine-tuning on existing VideoLMs.
Its effectiveness on larger and more diverse video-text data, such as longer videos, multi-event scenarios, and broader instruction-style tasks, is worth further exploration.
Since VT-Contrast provides auxiliary supervision by reshaping existing video-token representations, its gains may still depend on the temporal modeling and generation ability of the base model.
This is reflected by cases where larger models do not always perform better, such as TempCompass Caption, suggesting that the gains from VT-Contrast are affected by the base model's task-specific capability.
These observations point to a promising direction of incorporating temporal counterfactual supervision into larger-scale pretraining to further strengthen VideoLMs' temporal capability.

\section*{Acknowledgments}
This research is supported by the MOE AcRF Tier 1 Seed Grant (RS37/24, \#025041-00001), Singapore.

\bibliography{custom}

\clearpage
\appendix

\section{Additional Analysis}
\label{sec:appendix_exp}
In this appendix, we provide additional analysis of the proposed VT-Contrast.
We study the effects of the contrastive loss weight $\lambda$, targeted layer selection, the number $M$ of negative views, and the contrastive temperature $\tau_c$.
We also include qualitative examples to better illustrate the behavior of our model.
Overall, these results show that VT-Contrast is relatively robust to common hyperparameter choices.
Unless otherwise specified, we follow the main experimental setting and report results based on the 2B model with 8 input frames.

\subsection{Contrastive Loss Weight}
\label{sec:appendix_weight}

\begin{figure}[htbp]
    \centering
    \includegraphics[width=1\linewidth]{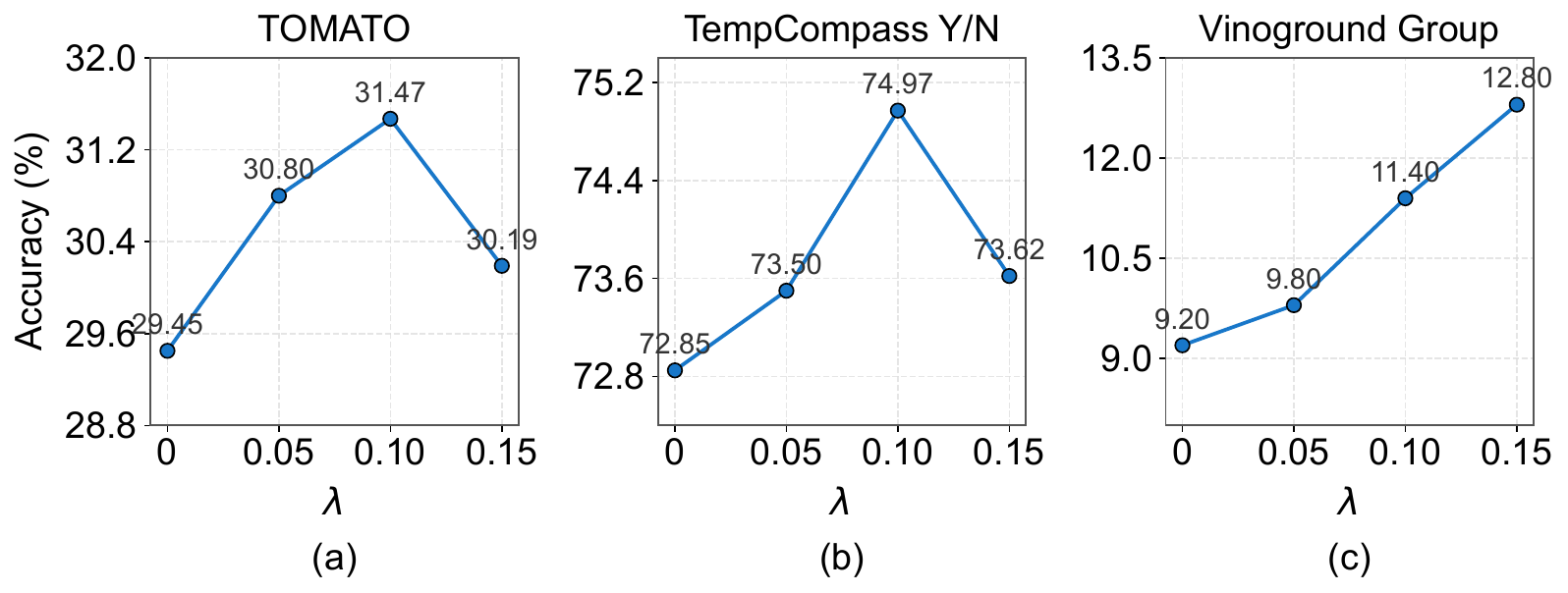}
\caption{
Effect of the contrastive loss weight $\lambda$.
Adding VT-Contrast generally improves performance over $\lambda=0$, while larger weights do not always help.
We use $\lambda=0.1$ as the default setting, which gives the best overall performance in this ablation.
}
    \label{fig:ab_weight}
\end{figure}

We first study the effect of the contrastive loss weight $\lambda$.
For TempCompass and Vinoground, we report TempCompass Y/N and Vinoground Group because the former is highly order-sensitive and the latter is the most representative Vinoground metric.
As shown in Figure~\ref{fig:ab_weight}, adding VT-Contrast improves performance over $\lambda=0$ on all three benchmarks.
As $\lambda$ increases from 0, the stronger contrastive signal generally improves temporal discrimination.
However, after a certain point, a larger weight can hurt some tasks, likely because the contrastive objective starts to interfere with the model's general downstream performance.
Therefore, we use $\lambda=0.1$ as the default setting, which gives the best overall trade-off in these results.

\subsection{Targeted Layer Selection}
\label{sec:appendix_layer}
\begin{figure}[htbp]
    \centering
    \includegraphics[width=1\linewidth]{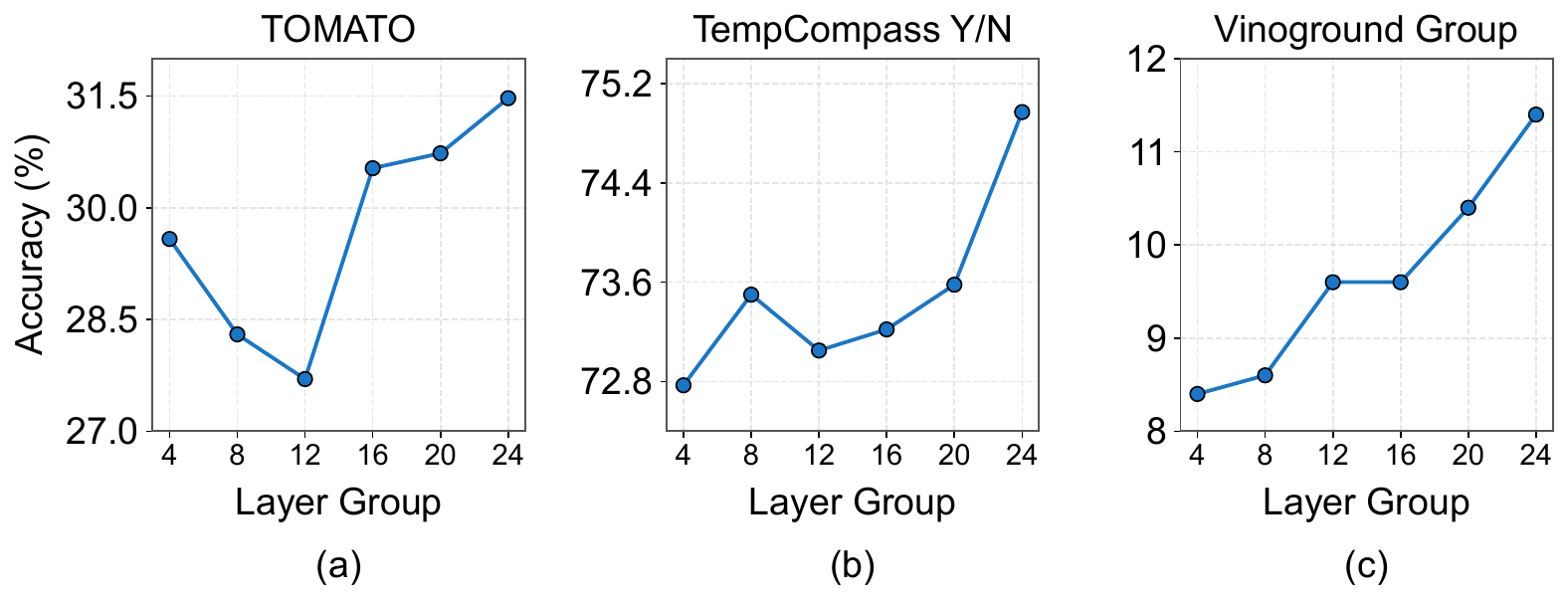}
\caption{
Effect of targeted layer selection.
Different layer ranges show different effects, highlighting the importance of targeted layer selection.
For the 2B model, we use layers 21--24, which perform best across the representative temporal tasks.
}
\label{fig:ab_layer}
\end{figure}

We further study contrastive layer selection using the same three representative tasks as in Appendix~\ref{sec:appendix_weight}.
As shown in Figure~\ref{fig:ab_layer}, different layer ranges lead to clearly different performance, indicating that where VT-Contrast is applied matters.
A general trend is that middle-to-late layers perform better than early layers.
This suggests that temporal information may be sufficiently aggregated in higher layers, while early-layer representations are still forming local visual features and are less suitable for our temporal counterfactual supervision.
These results highlight the importance of targeted layer selection for obtaining strong performance.
For the 2B model, we therefore apply VT-Contrast to layers 21--24.
Although the best layer range may vary across model scales and deserves further exploration, later layers generally provide a strong and reliable default choice.

\subsection{Number of Counterfactual Negatives}
\label{sec:appendix_num}

\begin{table*}[t]
\centering
\fontsize{8pt}{9pt}\selectfont
\renewcommand{\arraystretch}{1.1}
\begin{tabular}{
L{0.18\textwidth}
C{\metricw}
C{\metricw}
C{\metricw}
C{\metricw}
C{\metricw}
C{\metricw}
C{\metricw}
C{\metricw}
}
\toprule
\multirow{2}{*}{\textbf{Negative Number}}
& \multirow{2}{*}{\textbf{TOMATO}}
& \multicolumn{4}{c}{\textbf{TempCompass}}
& \multicolumn{3}{c}{\textbf{Vinoground}} \\
\cmidrule(lr){3-6} \cmidrule(lr){7-9}
&
& \textbf{Y/N}
& \textbf{MCQ}
& \textbf{Caption}
& \textbf{Match}
& \textbf{Text}
& \textbf{Video}
& \textbf{Group} \\
\midrule
4 samples
& 30.93
& \underline{73.79}
& \underline{69.56}
& 62.18
& 79.37
& 38.40
& 23.00
& \underline{10.80} \\

8 samples
& \textbf{32.21}
& 73.42
& 69.43
& \textbf{64.32}
& \underline{79.97}
& \underline{39.00}
& \underline{23.80}
& 10.20 \\

\oursrow
16 samples
& \underline{31.47}
& \textbf{74.97}
& \textbf{70.00}
& \underline{64.17}
& \textbf{80.17}
& \textbf{41.20}
& \textbf{25.00}
& \textbf{11.40} \\
\bottomrule
\end{tabular}
\caption{
Effect of the number $M$ of counterfactual negatives.
Using 4 negatives provides a relatively weak temporal counterfactual signal, while using 8 negatives already gives competitive results on several tasks.
Increasing to 16 negatives further improves overall stability and achieves the best overall performance.
}
\label{tab:ab_negative_num}
\end{table*}
\begin{table*}[t]
\centering
\fontsize{8pt}{9pt}\selectfont
\renewcommand{\arraystretch}{1.1}
\begin{tabular}{
L{0.18\textwidth}
C{\metricw}
C{\metricw}
C{\metricw}
C{\metricw}
C{\metricw}
C{\metricw}
C{\metricw}
C{\metricw}
}
\toprule
\multirow{2}{*}{\textbf{Temperature}}
& \multirow{2}{*}{\textbf{TOMATO}}
& \multicolumn{4}{c}{\textbf{TempCompass}}
& \multicolumn{3}{c}{\textbf{Vinoground}} \\
\cmidrule(lr){3-6} \cmidrule(lr){7-9}
&
& \textbf{Y/N}
& \textbf{MCQ}
& \textbf{Caption}
& \textbf{Match}
& \textbf{Text}
& \textbf{Video}
& \textbf{Group} \\
\midrule
0.06
& 29.99
& 73.58
& \textbf{70.51}
& \textbf{64.62}
& 79.24
& \textbf{43.60}
& 23.40
& 11.20 \\

0.08
& \underline{31.27}
& 73.54
& 69.68
& 63.67
& \textbf{80.17}
& 39.20
& \underline{24.60}
& \textbf{11.60} \\

\oursrow
0.10
& \textbf{31.47}
& \textbf{74.97}
& 70.00
& \underline{64.17}
& \textbf{80.17}
& \underline{41.20}
& \textbf{25.00}
& \underline{11.40} \\

0.12
& 30.39
& \underline{74.56}
& 69.68
& 62.57
& \underline{79.91}
& 40.60
& 23.40
& \underline{11.40} \\

0.14
& 29.65
& 73.05
& \underline{70.13}
& 63.72
& 79.31
& 40.60
& 22.60
& 10.80 \\
\bottomrule
\end{tabular}
\caption{
Sensitivity to the contrastive temperature $\tau_c$.
Different temperatures yield comparable results across most tasks, indicating that VT-Contrast is not highly sensitive to this hyperparameter.
We use $\tau_c=0.10$ by default, which provides a strong overall trade-off.
}
\label{tab:ab_temperature}
\end{table*}
\begin{figure*}[htbp]
    \centering
    \includegraphics[width=1\linewidth]{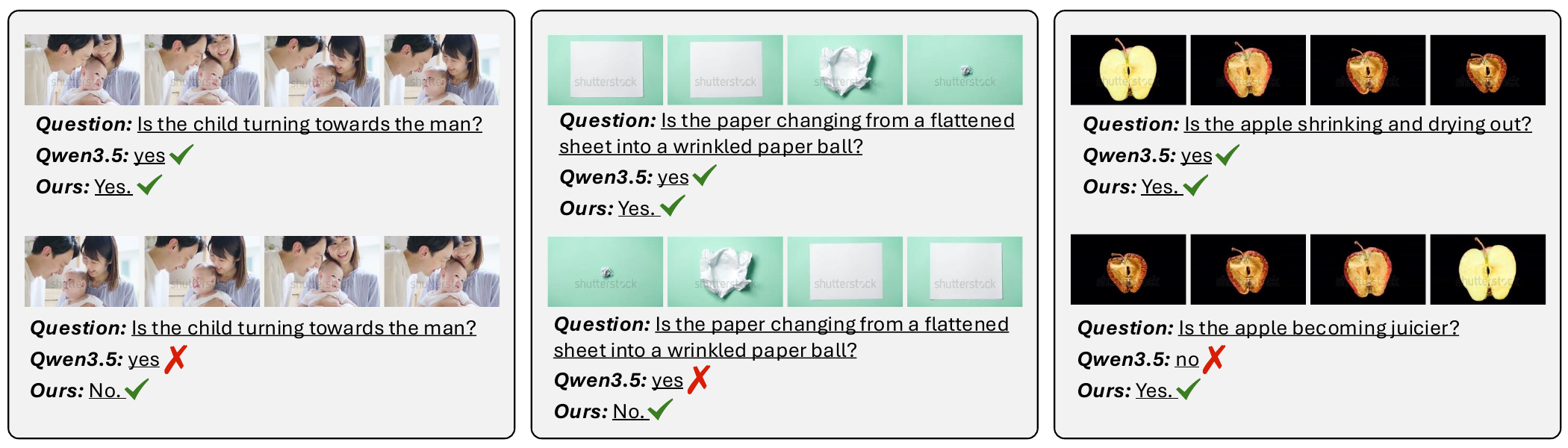}
\caption{
Qualitative examples on TempCompass Y/N with reversed video pairs.
Each example contains an original video and its reversed counterpart, which share similar visual content but require different answers.
Compared with Qwen3.5, our method better adapts its predictions to the changed temporal order.
}
    \label{fig:qualitative}
\end{figure*}

Table~\ref{tab:ab_negative_num} studies the number of temporal counterfactual negatives.
The number of negatives controls how many temporally perturbed same-video views are compared with each anchor in the InfoNCE objective.
With only 4 negatives, the contrastive signal is limited, leading to weaker overall performance.
Using 8 negatives already gives competitive results on several tasks, suggesting that moderate negative diversity is helpful.
Further increasing the number to 16 provides richer temporal counterfactual comparisons and leads to the most stable results.
Therefore, we use 16 counterfactual negatives as the default setting.

\subsection{Contrastive Temperature}
\label{sec:appendix_temperature}
Table~\ref{tab:ab_temperature} studies the sensitivity to the contrastive temperature $\tau_c$.
In the InfoNCE objective, $\tau_c$ controls the sharpness of the similarity distribution: smaller values make the objective more sensitive to similarity differences, while larger values produce smoother contrastive signals.
The results show that VT-Contrast is relatively robust to this hyperparameter, with different temperatures giving comparable performance across most tasks.
Among them, $\tau_c=0.10$ provides a strong trade-off, performing best on several tasks while remaining competitive on others.
A smaller temperature sharpens the contrastive distribution, which can help some tasks but also leads to less stable results.
By contrast, a larger temperature smooths the distribution and weakens the separation signal, leading to weaker overall performance.
We therefore use $\tau_c=0.10$ as the default setting.

\subsection{Qualitative Analysis}
\label{sec:appendix_qualitative}
Figure~\ref{fig:qualitative} presents qualitative examples from TempCompass Y/N, where each case contains an original video and its reversed counterpart.
We choose this setting because the original and reversed videos share nearly the same visual content, while the questions are designed to test whether the model can distinguish changes in event direction and temporal order.
As shown in the examples, Qwen3.5 is less sensitive to the temporal change between the original and reversed videos.
It fails on fine-grained temporal transitions, including the child's turning direction, paper deformation, and the apple's visual state change, while our method adapts more reliably to the reversed temporal semantics.
These cases suggest that VT-Contrast improves sensitivity to temporal progression and reduces reliance on static appearance cues.

\section{Additional Experiments}
\label{sec:appendix_experiments}
In this appendix, we provide additional experiments to further evaluate VT-Contrast beyond the main settings. We evaluate VT-Contrast with InternVL3.5~\cite{wang2025internvl3}, which belongs to a different VideoLM family, assess VT-Contrast on general video QA, and report the training overhead introduced by the contrastive objective.

\subsection{Evaluation with InternVL3.5}
To further assess the generalizability of VT-Contrast across different VideoLM families, we evaluate our method on InternVL3.5~\cite{wang2025internvl3}, beyond the Qwen3.5 model family considered in our main experiments.
Specifically, we use InternVL3.5-4B and train VT-Contrast with 8-frame inputs, while adapting the configuration by setting the contrastive loss weight to $\lambda=0.12$, applying the objective to layers 29--32, and reducing the number of negatives to $M=12$ for training efficiency.
As in our main experiments, we use a single model trained with 8-frame inputs and evaluate it at inference time using 8, 16, and 32 sampled frames.
Results in Table~\ref{tab:internvl_results} show that VT-Contrast improves most metrics across different frame settings, further supporting its effectiveness beyond the Qwen3.5 model family.
\begin{table*}[t]
\centering
\fontsize{8pt}{9pt}\selectfont
\setlength{\tabcolsep}{1.8pt}
\renewcommand{\arraystretch}{1.1}

\newcolumntype{F}{>{\centering\arraybackslash}p{0.06\textwidth}}
\newcolumntype{C}{>{\raggedright\arraybackslash}p{0.20\textwidth}}
\newcolumntype{A}{>{\centering\arraybackslash}p{0.05\textwidth}}
\newcolumntype{G}{>{\raggedright\arraybackslash}p{0.05\textwidth}}

\newcommand{\internposgain}[1]{%
{\hspace{-0.2em}\fontsize{5pt}{5pt}\selectfont\itshape
\textcolor{green!50!black}{#1}}}

\newcommand{\internneggain}[1]{%
{\hspace{-0.2em}\fontsize{5pt}{5pt}\selectfont\itshape
\textcolor{red!70!black}{#1}}}

\newcommand{\internzerogain}[1]{%
{\hspace{-0.2em}\fontsize{5pt}{5pt}\selectfont\itshape
\textcolor{gray!65!black}{#1}}}

\definecolor{internoursgray}{gray}{0.95}
\newcommand{\internoursrow}{\rowcolor{internoursgray}}

\begin{tabular}{
F C
A@{\hspace{0.0em}}G
A@{\hspace{0.0em}}G
A@{\hspace{0.0em}}G
A@{\hspace{0.0em}}G
A@{\hspace{0.0em}}G
A@{\hspace{0.0em}}G
}
\toprule
\multirow{2}{*}{\textbf{Frames}}
& \multirow{2}{*}{\textbf{Model}}
& {\multirow{2}{*}{\hspace{-0.83em}\textbf{TOMATO}}}
&
& \multicolumn{8}{c}{\textbf{TempCompass}}
& \multirow{2}{*}{\hspace{-0.9em}\shortstack[c]{\textbf{Vinoground}\\[-1pt]\textbf{Group}}} \\
\cmidrule(lr{12pt}){5-12}

&
&
&
& \textbf{Y/N} &
& \textbf{MCQ} &
& \makebox[0pt][l]{\hspace{-1.72em}\textbf{Caption}} &
& \textbf{Match} 
&& \\
\midrule

&
InternVL3.5-4B
& 28.71 &
& 72.08 &
& 68.73 &
& 60.93 &
& 79.44 &
& 16.00 & \\

\internoursrow
\cellcolor{white}\multirow{-2}{*}{\textbf{8}}
& Ours-4B
& \textbf{30.26} & \internposgain{+1.55}
& \textbf{72.40} & \internposgain{+0.32}
& \textbf{70.06} & \internposgain{+1.33}
& \textbf{63.67} & \internposgain{+2.74}
& 79.44 & \internzerogain{+0.00}
& \textbf{16.60} & \internposgain{+0.60} \\

\midrule

&
InternVL3.5-4B
& 30.12 &
& 72.89 &
& 70.70 &
& 61.68 &
& 80.44 &
& 16.80 & \\

\internoursrow
\cellcolor{white}\multirow{-2}{*}{\textbf{16}}
& Ours-4B
& \textbf{30.19} & \internposgain{+0.07}
& \textbf{73.38} & \internposgain{+0.49}
& \textbf{71.52} & \internposgain{+0.82}
& \textbf{64.32} & \internposgain{+2.64}
& 79.51 & \internneggain{-0.93}
& \textbf{18.20} & \internposgain{+1.40} \\

\midrule

&
InternVL3.5-4B
& 29.65 &
& 73.46 &
& 70.76 &
& 62.03 &
& 80.71 &
& 19.40 & \\

\internoursrow
\cellcolor{white}\multirow{-2}{*}{\textbf{32}}
& Ours-4B
& \textbf{30.05} & \internposgain{+0.40}
& \textbf{73.95} & \internposgain{+0.49}
& \textbf{71.90} & \internposgain{+1.14}
& \textbf{65.32} & \internposgain{+3.29}
& 80.71 & \internzerogain{+0.00}
& 19.40 & \internzerogain{+0.00} \\

\bottomrule
\end{tabular}

\caption{Results of VT-Contrast on InternVL3.5-4B under different frame settings. Small colored numbers indicate changes over the InternVL3.5-4B baseline. VT-Contrast improves most metrics across 8, 16, and 32 frames, further supporting its effectiveness beyond the Qwen3.5 model family.
}
\label{tab:internvl_results}
\end{table*}

\subsection{Evaluation on General Video QA}
We further evaluate VT-Contrast on general video QA to examine whether its benefits extend beyond temporally focused downstream tasks.
For this purpose, we adopt Video-MME~\cite{DBLP:conf/cvpr/FuDLLRZWZSZCLLZ25}, a comprehensive benchmark for evaluating general video understanding.
We use Qwen3.5-2B as the base model and train it for 1,500 steps on a subset of the open-ended QA data from LLaVA-Video-178K~\cite{zhang2024llava}. 
Compared with the action-centric SSv2 data used in our main experiments, this setting provides more general supervision, allowing us to assess whether VT-Contrast also benefits broader video understanding.
Results in Table~\ref{tab:videomme} show that our method improves the overall Video-MME accuracy.
These results further demonstrate the applicability of VT-Contrast to broader video QA settings.
\begin{table}[t]
\centering
\fontsize{8pt}{9pt}\selectfont
\renewcommand{\arraystretch}{1.1}
\begin{tabular}{
L{0.28\columnwidth}
C{0.16\columnwidth}
C{0.16\columnwidth}
C{0.16\columnwidth}
}
\toprule
\multirow{2}{*}{\textbf{Method}}
& \multicolumn{3}{c}{\textbf{Frames}} \\
\cmidrule(lr){2-4}
& \textbf{8}
& \textbf{16}
& \textbf{32} \\
\midrule

Qwen3.5-2B
& 52.07
& 54.93
& 57.85 \\

Ours-2B
& \textbf{52.67}
& \textbf{56.07}
& \textbf{58.30} \\

\oursrow
{\gain{Gain}}
& \gain{+0.60}
& \gain{+1.14}
& \gain{+0.45} \\

\bottomrule
\end{tabular}

\caption{
Results on general video QA with Video-MME, reporting overall accuracy. Gains denote absolute improvements over the Qwen3.5-2B baseline, showing the effectiveness of our method for general video QA.
}
\label{tab:videomme}
\end{table}

\subsection{Training Overhead}
We additionally report the training overhead introduced by VT-Contrast.
Since each training sample includes an anchor view, a positive view, and multiple counterfactual negatives, the method requires additional computation compared with standard training.
We profile Qwen3.5 models on an 8-H100-GPU server with a per-device batch size of 1 and a fixed effective global batch size of 16, comparing $M=0,4,8,16$ negatives.
For each configuration, we conduct five 40-step runs, discard the first 10 steps of each run as warm-up, compute statistics over the remaining 30 steps, and report the average across the five runs.
As shown in Table~\ref{tab:training_overhead}, increasing $M$ introduces additional training overhead as expected.
These results characterize the practical training cost of VT-Contrast.
Since our current implementation is not specifically optimized for efficiency, further improvements may be possible with additional optimization.
\begin{table}[t]
\centering
\fontsize{8pt}{9pt}\selectfont
\renewcommand{\arraystretch}{1.1}
\setlength{\tabcolsep}{3.5pt}
\begin{tabular}{
@{}
L{0.095\columnwidth}
L{0.235\columnwidth}
C{0.125\columnwidth}
C{0.125\columnwidth}
C{0.125\columnwidth}
C{0.125\columnwidth}
@{}
}
\toprule
\textbf{Model}
& \textbf{Metric}
& \multicolumn{4}{c}{\textbf{Number of Negatives $M$}} \\
\cmidrule(lr){3-6}
&
& \textbf{0}
& \textbf{4}
& \textbf{8}
& \textbf{16} \\
\midrule

\multirow{3}{*}{0.8B}
& Time (s)
& 1.83 & 5.09 & 5.26 & 6.28 \\
& Alloc. (GiB)
& 9.73 & 11.35 & 11.90 & 13.86 \\
& Reserved (GiB)
& 15.32 & 16.44 & 17.28 & 19.23 \\
\midrule

\multirow{3}{*}{2B}
& Time (s)
& 1.94 & 5.42 & 5.75 & 7.09 \\
& Alloc. (GiB)
& 24.84 & 26.54 & 26.94 & 29.56 \\
& Reserved (GiB)
& 39.53 & 40.86 & 42.88 & 45.46 \\
\midrule

\multirow{3}{*}{4B}
& Time (s)
& 2.57 & 7.05 & 8.18 & 10.77 \\
& Alloc. (GiB)
& 51.36 & 52.38 & 53.52 & 57.39 \\
& Reserved (GiB)
& 70.88 & 72.82 & 74.99 & 77.40 \\
\bottomrule
\end{tabular}

\caption{
Training overhead of VT-Contrast. $M=0$ denotes standard training without the contrastive objective. Time denotes wall-clock time per training step, while Alloc. and Reserved denote peak allocated and reserved CUDA memory, respectively.
}
\label{tab:training_overhead}
\end{table}

\section{Prompt Details}
\label{sec:appendix_prompt}
\newtcolorbox{promptbox}{
  enhanced,
  breakable,
  colback=gray!3,
  colframe=black!85,
  boxrule=0.45pt,
  arc=2pt,
  left=6pt,
  right=6pt,
  top=5pt,
  bottom=5pt,
  before skip=6pt,
  after skip=6pt,
  fontupper=\fontsize{9pt}{11pt}\selectfont\itshape
}

For reproducibility, we list the prompts used for benchmark evaluation.
Across all reported models and settings, we use the same prompt format within each benchmark to reduce prompt-induced variation.
For each task, we present a prompt example illustrating the input format.

\subsection{TOMATO}

TOMATO is evaluated as a multiple-choice task.
Following the default prompt in the lmms-eval framework, we further constrain the model to answer with a single option letter.
This reduces overly open-ended responses and simplifies the subsequent evaluation process in practice.
An example prompt is shown below.
\begin{promptbox}
You will be provided with $n$ separate frames uniformly sampled from a video, the frames are provided in chronological order of the video. Analyze these frames and provide the answer to the question about the video content. Answer the multiple-choice question about the video content.

\vspace{9pt}
You must use these frames to answer the multiple-choice question; do not rely on any externel knowledge or commonsense. 

\vspace{9pt}
<question> 

In which direction(s) did the person's hand move? 

</question>

\vspace{9pt}
<options> 

{`A': `Not moving at all', `B': `Left.', `C': `Right.', `D': `First to the right then to the left.', `E': `First to the left then to the right.'}

</options>

\vspace{9pt}
Even if the information in these separate frames is not enough to answer the question, PLEASE TRY YOUR BEST TO GUESS AN ANSWER WHICH YOU THINK WOULD BE THE MOST POSSIBLE ONE BASED ON THE QUESTION. 

\vspace{9pt}
DO NOT GENERATE ANSWER SUCH AS `NOT POSSIBLE TO DETERMINE.'

\vspace{9pt}
Your output must be exactly one uppercase option letter from A, B, C, D, E.

\vspace{9pt}
Answer:
\end{promptbox}

\subsection{TempCompass}

For TempCompass, we use task-specific prompts for Yes or No, Multi Choice, Captioning, and Caption Matching.
As with TOMATO, we explicitly specify the required answer format when applicable to ensure consistent evaluation across models.
For Captioning, we follow the lmms-eval default and use GPT-3.5-Turbo for GPT-based grading.
Example prompts are provided below.
\begin{promptbox}
\textbf{Yes or No}:

Is the man dunking?

Please answer yes or no:

\vspace{9pt}
\textbf{Multi Choice}:

What is the man doing in the video?

A. dunking a basketball

B. dribbling a basketball

C. passing a basketball

Please only give the letter of the best option from A, B, C:

\vspace{9pt}
\textbf{Captioning}:

You will be presented with a video and several pieces of information. One piece of information is consistent with the video while the others are not. Please identify the information that consistent with the video and generate a video caption accordingly.

Information A: {`subject': `man', `action': `dribbling a basketball'}

Information B: {`subject': `man', `action': `dunking a basketball'}

Information C: {`subject': `man', `action': `passing a basketball'}

Generated Caption:

\vspace{9pt}
\textbf{Caption Matching}:

Which description is a more suitable match for the video?

Option 1: The man is dribbling a basketball.

Option 2: A man is dunking a basketball.

Please only give the label of the best option from Option 1, Option 2:
\end{promptbox}

\subsection{Vinoground}

For Vinoground, we use separate prompts for text-to-video and video-to-text matching.
Depending on the matching direction, the model chooses between two candidate videos or captions and outputs only the corresponding option letter.
Example prompts are shown below.
\begin{promptbox}
\textbf{Text:}

Which caption best describes this video?

A. a toddler picks up a water bottle and drinks before he plays around the grass field

B. a toddler plays around the grass field before he picks up a water bottle and drinks

Answer with the option's letter from the given choices directly.
Please only output one English character.

\vspace{9pt}
\textbf{Video}:

Which video segment matches this caption? Note: The video contains two segments separated by a 2-second black frame. Caption: a toddler plays around the grass field before he picks up a water bottle and drinks

A. First segment (before black frame)

B. Second segment (after black frame)

Answer with the option's letter from the given choices directly.
Please only output one English character.
\end{promptbox}


\end{document}